# Hierarchical Conflict Propagation: Sequence Learning in a Recurrent Deep Neural Network

Andrew J.R. Simpson [#1]

[#] *Centre for Vision, Speech and Signal Processing, University of Surrey*
*Surrey, UK*
[1] `Andrew.Simpson@surrey.ac.uk`

*Abstract*—Recurrent neural networks (RNN) are capable of learning to encode and exploit activation history over an arbitrary timescale. However, in practice, state of the art gradient descent based training methods are known to suffer from difficulties in learning long term dependencies. Here, we describe a novel training method that involves concurrent parallel cloned networks, each sharing the same weights, each trained at different stimulus phase and each maintaining independent activation histories. Training proceeds by recursively performing batch-updates over the parallel clones as activation history is progressively increased. This allows conflicts to propagate hierarchically from short-term contexts towards longer-term contexts until they are resolved. We illustrate the *parallel clones* method and hierarchical conflict propagation with a character-level deep RNN tasked with memorizing a paragraph of Moby Dick (by Herman Melville).

## I. INTRODUCTION

In principle, recurrent neural networks (RNN) are powerful general computing machines capable of learning long-term dependencies in sequences [1]. However, in practice, traction in supervised learning problems has been limited by difficulties in the optimization problem; Gradient descent based training methods appear insufficiently powerful to learn long-term dependencies [2] and this is thought to be due to the so-called vanishing gradient problem [3,4].

Intuitively, long-term dependencies are problematic because RNN sequence learning proceeds one step at a time. At each step, only a given input and a given output are available for training, so the weights of the RNN are optimized to reduce single-step prediction error at the output. Consider a character-level model, within a next-character-prediction paradigm, tasked with learning the sequence of seven characters in the string "`ace act`". This yields a dictionary of ["`a`", "`c`", "`e`", "`t`", " "] which may be encoded in one-hot binary form (e.g., "`a`" = [1 0 0 0 0]). A large part of the problem can be solved with only a feed-forward network architecture (i.e., which is agnostic to history) to provide the following unique mappings; "`a`"→"`c`", "`e`"→" " and " "→"`a`". From a gradient descent point of view, these mappings are straight forward to solve because the respective weights may be set as mutually exclusive mappings. However, the mappings "`c`"→"`e`" and "`c`"→"`t`" – occurring in the first and second words - are conflicting. Therefore, these two mutually exclusive mappings represent a problem for gradient descent. Using an RNN, the problem may be solved in principle by storing the activation history so that it may later be used to disambiguate conflicts at the immediate (i.e., feed-forward) level. This process of storage is known as 'latching' [3]. Specifically, if the earlier activation of "`a`" or "`c`" or "`e`" or " " is latched then it may be combined with the input of "`c`" (in the second word "`act`") in order to predict a "`t`" (instead of the conflicting "`e`" of "`ace`").

This example illustrates the problem of learning to resolve short-term conflicts according to some longer-term contextual information. Extrapolating, if such conflicts also arise in the longer-term contexts, then we must exploit yet longer-term contextual information in order to resolve the conflict. Thus, in principle, conflicts must propagate through a hierarchy, from short-term towards long-term, until resolution. Hence, it is not surprising that linear (sequential) gradient descent methods (e.g., online or back propagation through time [5-7]) do not provide a good solution.

In this paper, we describe a novel *parallel clones* method for training deep RNN (DRNN) according to the principle of hierarchical conflict propagation. Our method embodies the principle that unresolvable conflicts may be propagated from the short-term historical context towards the long-term historical context until they are resolved. Our method involves a number of parallel DRNNs, each an identical clone of the *target* DRNN. Each parallel clone shares the same set of weights but maintains independent activation history by operating at independent phase (position) within the training sequence.

Each complete iteration of training features a full sweep of the training sequence. Each of the clones begins at a different point in the sequence and the sweep proceeds in a circular fashion. At each step of the circular sweep, weight update gradients are computed and averaged over the parallel clones before being applied (in an online fashion). This means that each batch update is averaged over the entire sequence and that only the histories progressively diverge as the sweep progresses. This whole-sequence batch averaging allows unresolved conflicts to be propagated towards resolution at longer-term contexts. We capture the hierarchical propagation of conflict by visualising the evolving distribution of the loss function over different degrees of history as training progresses. This allows us to characterise the shifting of loss from long-term contexts to short-term contexts. We demonstrate the method by using it to train a character-level DRNN to memorize the first 500 characters of Moby Dick (the book by Herman Melville).

## II. METHOD

We consider a typical next-character prediction paradigm featuring the first 500 characters of the opening paragraph of the book 'Moby Dick' by Herman Melville. Figure 1 provides a verbatim account of the text. This excerpt contains 42 unique characters and hence requires a dictionary of length 42. Each character was encoded in a one-hot (or, 1-in-*k*) binary vector of length 42, where the corresponding dictionary entry was set to the value of one and the rest of the vector was set to zero. Thus, the entire training sequence provided a matrix of encoded characters of size 42 (dictionary length) by 500 (number of characters).

Figure 1 provides an illustrated account of the exact text of the training sequence, including formatting resulting from formatting characters (i.e., line returns and spaces). There are several obvious conflicts at various levels of the temporal hierarchy (as highlighted in various colours). For example, at the first (i.e., historical context) level there is "`ne`", "`nd`", "`ng`", "`no`", etc. At the fourth level there is "`and n`" and "`and s`" and "`and r`". Thus, it appears necessary to propagate conflicts involving predictions for "`n`" prior to resolving conflicts involving "`and` ". At the fifth level there is "`in my p`[urse]" and "`in my s`[oul]". Therefore, intuitively from Fig. 1, there is an obvious need for conflict propagation over a temporal hierarchy.

```
"CHAPTER 1. Loomings.

Call me Ishmael. Some years ago--never mind how long precisely--having
little or no money in my purse, and nothing particular to interest me on
shore, I thought I would sail about a little and see the watery part of
the world. It is a way I have of driving off the spleen and regulating
the circulation. Whenever I find myself growing grim about the mouth;
whenever it is a damp, drizzly November in my soul; whenever I find
myself involuntarily pausing before coffin warehouses, and br"
```

*First level conflicts*   : "`me`", "`ma`", "`mi`", "`m `", "`mo`", "`my`", "`mp`", "`mb`"
*Second level conflicts* : "`way`", "`wat`", "`no `", "`not`"
*Fourth level conflicts* : "`and n`", "`and s`", "`and r`","`and b`",
*Fifth level conflicts*   : "`in my p`", "`in my s`",

**Fig. 1. Moby Dick text training sequence with selected conflicts illustrated.** The RDNN was tasked with learning a character-level model of the above opening excerpt from Moby Dick by Herman Melville. Like colours indicate conflicts at different levels of temporal hierarchy.

We constructed a recurrent deep neural network with input layer of dimension 42 + 256 + 42 = 340, hidden layer of dimension 256, and output layer of dimension 42 (representing the one-hot encoding of a character). The input layer was a concatenation of the input vector (i.e., input character: vector of length 42) and the hidden layer activations (length 256) and output layer activations (predicted character: vector of length 42) at the previous time step. I.e., this made the network recurrent. Biased sigmoid activation functions [8] were used with a *softmax* output layer. This constituted the *target* RDNN. A generalised schematic diagram of the *target* RDNN, showing feed-forward and recurrent connections, is given in Figure 2. A learning rate of 1 was used throughout.

The *target* RDNN was initialised with random weights and then subsequently cloned *N* times (*N*=499). Each of the *N* clones ($C_n$) addressed the training sequence ($S$) from a different (*n*th) location at any given moment (i.e., they each indexed the training sequence at a different phase). All the *N* clones swept the training sequence in parallel, calculating weight updates using backpropagation gradient descent [6] for each parallel clone at each step of the sweep, averaging the weight updates (gradients) and applying the averaged update to the shared weights after each step.

The *n*th clone ($C_n$) swept the training sequence in a circular fashion with sweep index ($q$) proceeding from $q$=1 to $q$=499 in steps of 1. Activation history (i.e., the recurrent activations fed into the input layer) was zeroed for the first step at $q$=1. The *n*th clone then calculated the weight update to minimize error for prediction of the character at $S_{1+mod(-2+n+q+1,500)}$ from the character at $S_{1+mod(-2+n+q,500)}$. Thus, for every batch-averaged update, the entire sequence was considered. This means that, during the full sweep, all possible historical contexts were considered (from zero history to the maximum history for each possible index into the training sequence).

For comparison, the *target* RDNN was replicated (with identical random weight initialisation) and was trained using a basic online gradient descent, where each full sweep of the training sequence proceeded from the beginning to the end with updates after each step. We will call this the *regular* RDNN. In addition, an equivalent 499 *non-active* clones were obtained each sharing the weights of the *regular* RDNN and each sweeping across the training sequence at different phase exactly as described for the (*active*) parallel clones used to train the *target* RDNN. The only difference was that these clones were not involved in training (i.e., they were *non-active*) but were only used to compute the loss function across the historical contexts (across the clones) for comparison.

At each step of the sweep, the cross entropy loss function was evaluated over the entire training sequence by evaluating the feed-forward predictions of each of the clones at that step (since the whole set of clones combined address the entire sequence at any given moment) and taking the mean of all these cross entropy loss measures. This was done for both the *target* RDNN (using the parallel clones) and with the *regular* RDNN (using the *non-active* parallel clones). This allows us to track the propagation of loss across historical context as training progresses, allowing us to compare the two learning methods so as to identify evidence to support a claim of conflict propagation. For both *target* and *regular* models, training was conducted over 100 full-sweep iterations. In addition, the sum of the mean-loss functions, over all historical points, was obtained for each full-sweep training step. This allows us to identify whether or not the overall learning was monotonic across history and hence interpret the history-dependent loss functions in terms of shifts in the distribution across history.

In addition to the analysis of the loss function, the *target* and *regular* models were also tested for recall accuracy after 100 full-sweep iterations of training. To do this, each network was fed (sequentially, at the input) with the first 10 characters of the training sequence. After this 10-character 'seeding', each network was then fed its output prediction as the input and allowed to continue (i.e., without external support) for the remaining 490 characters. Then, the output stream was decoded using the dictionary and recall accuracy was evaluated using the Levenshtein edit distance metric [9], which captures the degree of editing necessary to correct the predicted text to match the training sequence.

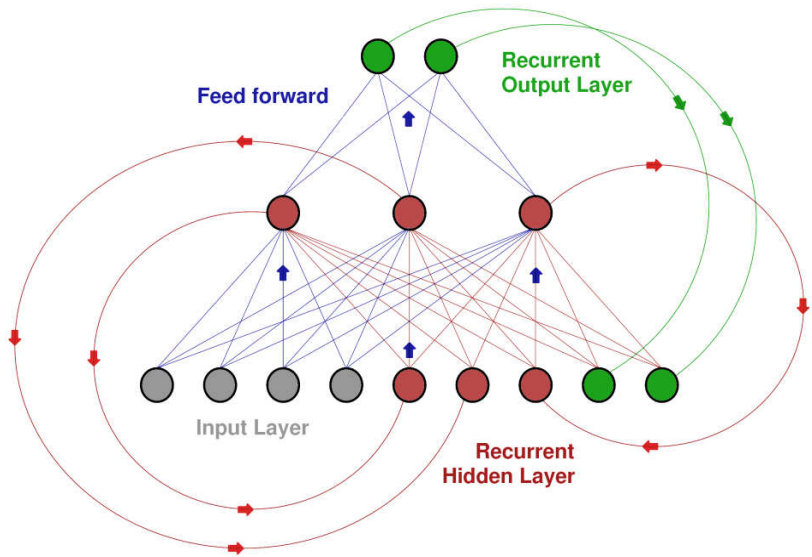

**Fig. 2. Recurrent deep neural network - schematic diagram.** The *target* and *regular* RDNNs (and their clones) feature recurrent connections such that the activations in the hidden and output layer (at the previous time step) are fed back into the input layer.

## III. RESULTS

Figure 3 plots the various cross entropy loss functions for the *regular* and *target* models. Figs. 3a and 3b plot the historical-context-dependent loss functions for the *regular* and *target* models respectively. Only the first 50 historical contexts are plotted for clarity as the remaining longer-term historical contexts follow (in an exponential fashion) the same trend. Fig. 3c plots the respective functions showing the sums-over-historical-contexts (at each full-sweep training iteration point) for the *regular* and *target* models. Overall, in Fig. 3c, the losses are much smaller for the *target* model, indicating that the parallel clones training is much more effective than the regular online gradient descent. In Figs. 3a and 3b, the contrast is even greater for longer historical levels (shown in red), where loss falls rapidly for the *target* model indicating the effective learning of long-term dependencies. Also, the *regular* functions are noisy/irregular whereas the *target* functions are smooth.

*Conflict propagation.* In both the *regular* (Fig. 3a) and *target* (Fig. 3b) history-dependent-loss plots, the functions at different historical levels are initially similar or equivalent, indicating a near-uniform distribution of loss over the various historical contexts. However, the loss functions diverge as training progresses. In particular, where there is the least history (blue) the functions are non-monotonic in both cases (*regular* and *target*). This means that loss in the short-historical contexts *increased* with training whilst the sum loss (over all historical contexts – plotted in Fig. 3c) monotonically *decreased*.

In the case of the *target* model (Fig. 3b), the non-monotonic functions are more defined and more abrupt and appear to follow a clear progression – the onset (i.e., the up-swing in the loss function) occurs progressively later (in full-sweep iteration time) as historical context is increased. This can be interpreted as evidence of conflict (i.e., represented as loss) propagation through the temporal-historical hierarchy. By contrast, such trends are not readily discernible in the *regular* functions (Fig. 3a), either because the functions are sufficiently noisy to mask such details or simply because the propagation is not pronounced enough to be obvious (or even perhaps it does not occur). Finally, in the case of the *regular* model, only the first two or three historical levels (i.e., zero history and 1 or 2 steps of history) show clear evidence of non-monotonic loss functions, indicating limited conflict propagation in this case. By contrast, at least the first 10-to-15 historical levels in the *target* loss functions show clear non-monotonic trends (Fig. 3b), indicating farther propagation of conflicts than was achieved with the *regular* model (Fig. 3a).

General to both models, over the first 50 historical levels plotted in Fig. 3, ultimate loss (at iteration 100) appears to be inversely proportional to historical level; for the *target* model this trend is significant ($r = -0.97$, $P < 0.001$, *Spearman rank correlation*), but for the *regular* model the trend is not significant ($r = 0.05$, $P > 0.1$, *Spearman rank correlation*). Therefore, in the case of the *target* model, this confirms the propagation of loss towards a distribution which is proportional to historical context, and tends to provide further evidence of conflict propagation over the temporal hierarchy.

After 100 full-sweep iterations of training, the Levenshtein edit distance was still at 407 (maximum possible edit distance being 500) for the *regular* RDNN. By contrast, after the same 100 full-sweep iterations of parallel-clones training, the edit distance had reached zero for the *target* RDNN (zero indicating perfect procedural recall). We do not show the output (prediction) of the *target* model here because it is identical to the training sequence. However, the erroneous output of the *regular* model (including the seed and after decoding via the dictionary) was:

"[new line] `APTER W. Loomo`",

followed by a continuous stream of 'blank space' characters (" ") until the 500 character limit. This indicates that the *regular* model entirely failed to replicate the sequence for more than 5 steps beyond the 10-character seed (and, indeed, made a mistake with the chapter number "`W`"). Taken together, it is clear that the *regular* model was not able to propagate conflicts in order to capture longer-term dependencies, whereas the *target* model (trained with the parallel clones method) was able to capture the entire sequence through effective hierarchical conflict propagation.

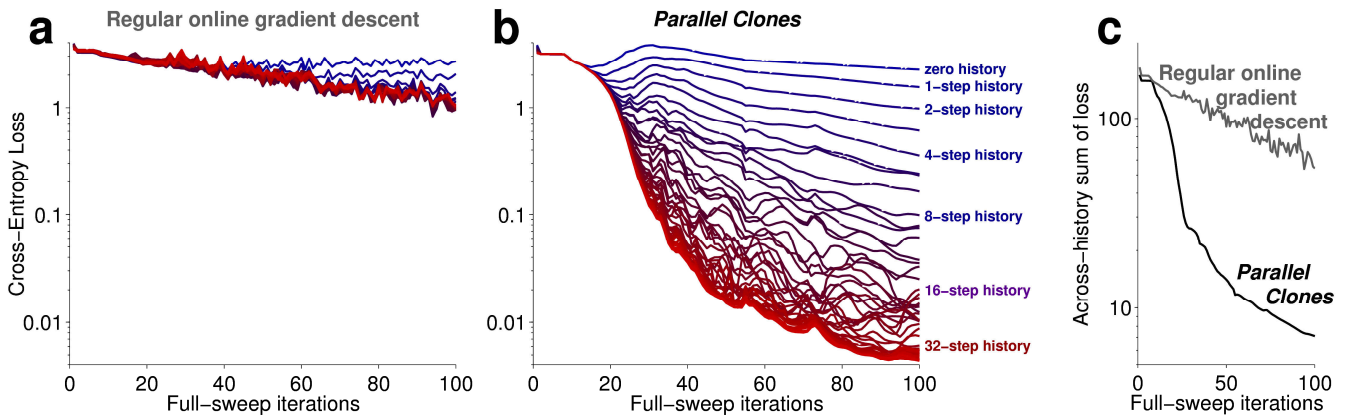

**Fig. 3. Cross entropy loss functions of full-sweep iterations for different historical contexts.** The parallel clones (both *active* clones, in the case of the *target* model and *non-active* clones in the case of the *regular* model) were used to compute mean cross entropy loss across the entire training sequence at each historical context level as training progressed. **a** plots the historical loss functions for the *regular* model and **b** plots the respective historical loss functions for the *target* model (trained with the parallel clones). Only the first 50 historical levels are shown. Note the logarithmic vertical axes (for the loss). Colours indicate historical level (i.e., how many steps of activation history were available) from blue (zero history) to red (50 steps of history). **c** plots the sum of the mean cross entropy loss functions over the different historical levels for each training iteration for the *regular* and *target* models respectively. These functions allow us to gauge whether the total loss functions were monotonic (thus allowing us to interpret non-monotonic functions in panels **a** and **b** as depicting shifts in the distribution of loss).

IV. DISCUSSION AND CONCLUSION

In this paper we have described a novel method for training recurrent deep neural networks to learn sequences and have illustrated the method with a procedural learning problem concerning next-character prediction using an excerpt from a popular work of fiction. We have argued that conflicts are propagated hierarchically from short-term historical contexts towards longer-term historical contexts. We have captured the hierarchical propagation of conflict according to the loss functions at different levels of historical context; loss is shifted towards earlier historical contexts as increasingly long-term contexts are employed in conflict resolution.

In this study, we chose not to make the issue of interpretation more difficult by conflating the learning problem with the distinctly separate *generalisation* problem.

Indeed, there is no reason to presume that the learning problem (i.e., of vanishing gradients, or of conflict propagation) has any meaningful bearing on, or relation to the problem of generalisation of learning in RNN. Nor did we wish to conflate the issues of learning sequences with those of learning in continuous feature spaces (e.g., for image recognition). However, in principle there is no reason that the concept of conflict propagation should not apply to such cases. Indeed, anecdotally, the method performs similarly for learning generalized RNN models which operate on continuous data (i.e., in a continuous abstract feature space) to perform classification or synthesis (data not shown). In brief, we implemented a similar architecture which allocated a set of independent parallel clones to each of a batch-super-set of training examples. I.e., for a batch size of $M$, there were $M$ sets of $N$ parallel clones. Batches were selected randomly (as in stochastic gradient descent), then, for each batch, we then swept over the $M$ examples in the manner described above, averaging the updates over the batch of $M$x$N$ parallel clones in the same circular, online manner.

The method described here was illustrated using circular indexing because it is the most simple and complete configuration of the method with regards to the representation of history. However, anecdotally, the parallel clones method works equally well when applied to non-circular indexing with minor modifications (data not shown). In addition, the method described here also works equally well (e.g., in the present test case) when fewer parallel models are applied (e.g., at spaced intervals throughout the training sequence). Finally, The method is also applicable to deep recurrent networks trained to learn continuous data (e.g., images or audio), where the intuitions regarding conflicts may be interpreted in terms of 'demodulation conflicts' in abstract feature space. Thus, the method described here may be instantiated in a number of possible configurations without departing substantially from the spirit and scope of the method as described here.

At this stage it is unclear as to how these results might be interpreted in terms of the vanishing gradient problem. However, the evidence of only limited short-term conflict propagation for the *regular* model (Fig. 3a) – trained with online gradient descent - is consistent with the anticipated result of a vanishing gradient problem (only short-term dependencies are learned). By contrast, the *target* model (trained with the parallel clones method) does not appear to have suffered from vanishing gradients.

Finally, an obvious strength of the parallel clones method is that it is inherently parallel and hence the method is suitable for efficient implementation over distributed computing architectures (e.g., multi-core processors and/or GPUs).

ACKNOWLEDGMENT

AJRS did this work on the weekends and was supported by his wife and children.